\documentclass[conference]{IEEEtran}
\usepackage{times}

\usepackage[numbers]{natbib}
\usepackage{graphicx}
\usepackage{multicol}
\usepackage{xcolor}
\usepackage{booktabs}
\usepackage{tikz}
\usepackage{amsthm}
\usepackage{balance}

\usepackage{freytex}
\newcommand*\x{{\vc{x}}}
\renewcommand*\u{\vc{u}}
\newcommand*\w{\vc{w}}
\newcommand*\Xsafe{{\manX_{\text{safe}}}}

\newcommand*\cXsafe{\manX_{\text{safe}}^c}
\newcommand*\tc[1][\,]{{^{#1\!}\tau}}
\newcommand*\Fz[1][{}]{F_{{#1}}}
\newcommand*\Fhat{{\hat{F}}}
\newcommand*\Phiz[1][{}]{\Phi_{{#1}}}
\newcommand*\pT{\mc{T}}  
\newcommand*\cbar{\,|\,}

\newcommand*\Ft[1][t]{\mc{F}_{#1}}  
\newcommand*\setOm{{\rm{O}^-}}  

\newcommand*\z{\vc{z}}
\newcommand*\zn[1][n]{\z^{(#1)}}
\newcommand*\zjn[1][j]{z_{#1}^{(n)}}

\newcommand*\y{\vc{y}}
\newcommand*\yhat{\hat{\vc{y}}}
\newcommand*\yjhat[1][j]{\hat{y}_{#1}}
\newcommand*\yhatn{{\yhat^{(n)}}}

\newcommand*\deltan{{\delta^{(n)}}}
\newcommand*\epsn{{\eps^{(n)}}}
\newcommand*\en{{e^{(n)}}}
\newcommand*\Kn{{K^{(n)}}}
\newcommand*\pTn{{\pT^{(n)}}}

\newcommand*\h{{\vc{h}}}
\newcommand*\f{{\vc{f}}}
\newcommand*\G{{\mx{G}}}

\newcommand*\p{\vc{p}}

\newcommand*\Rnp{{\R^{n_p}}}

\newcommand*\Sn[1][(n)]{S^{#1}}
\newcommand*\An[1][(n)]{A^{#1}}
\newcommand*\Bn[1][(n)]{B^{#1}}
\newcommand*\Cn[1][(n)]{C^{#1}}
\newcommand*\Dn[1][(n)]{D^{#1}}
\newcommand*\En[1][(n)]{E^{#1}}

\newcommand*\Snc{\Sn[(n),c]}
\newcommand*\Bnc{\Bn[(n),c]}
\newcommand*\Enc{\En[(n),c]}

\graphicspath{{figures/}{./}}

\newcommand*{\ARXIVVERSION}{}

\ifdefined\ARXIVVERSION
  \newcommand*\refAppProofs{Appendix \ref{app:proofs}}
  \newcommand*\refAppRedQuad{Appendix \ref{app:red_quad}}
  \newcommand*\refAppSys{Appendix \ref{app:sys}}

  \newcommand*\href[2]{#2}  
\else
  \usepackage[bookmarks=true]{hyperref}

  \newcommand*\refSupp[1][{}]{the supplementary material \cite[#1]{suppmat}}
  \newcommand*\refAppProofs{\refSupp[Appendix A]}
  \newcommand*\refAppRedQuad{\refSupp[Appendix B]}
  \newcommand*\refAppSys{\refSupp[Appendix C]}
\fi

\begin{document}

\ifdefined\ARXIVVERSION
  \title{Collision Probabilities for Continuous-Time Systems\\Without Sampling\\
  \Large{[with Appendices]}}
\else
  \title{Collision Probabilities for Continuous-Time Systems\\Without Sampling}
\fi


\author{
  \IEEEauthorblockN{Kristoffer M.\ Frey\IEEEauthorrefmark{1}\IEEEauthorrefmark{2}, Ted J.\ Steiner\IEEEauthorrefmark{2}, and Jonathan P.\ How\IEEEauthorrefmark{1}}
  \IEEEauthorblockA{
    \IEEEauthorrefmark{1}Department of Aeronautics and Astronautics, MIT\\
    \IEEEauthorrefmark{2}The Charles Stark Draper Laboratory, Inc.\\
    Cambridge, MA 02139\\
    Email: kfrey@mit.edu
  }
}

\maketitle

\begin{abstract}
Demand for high-performance, robust, and safe autonomous systems has grown substantially in recent years.
These objectives motivate the desire for efficient safety-theoretic reasoning that can be embedded in core decision-making tasks such as motion planning, particularly in constrained environments.
On one hand, Monte-Carlo (MC) and other sampling-based techniques provide accurate collision probability estimates for a wide variety of motion models but are cumbersome in the context of continuous optimization.
On the other, ``direct'' approximations aim to compute (or upper-bound) the failure probability as a smooth function of the decision variables, and thus are convenient for optimization.
However, existing direct approaches fundamentally assume discrete-time dynamics and can perform unpredictably when applied to continuous-time systems ubiquitous in the real world, often manifesting as severe conservatism.
State-of-the-art attempts to address this within a conventional discrete-time framework require additional Gaussianity approximations that ultimately produce inconsistency of their own.
In this paper we take a fundamentally different approach, deriving a risk approximation framework directly in continuous time and producing a lightweight estimate that actually converges as the underlying discretization is refined.
Our approximation is shown to significantly outperform state-of-the-art techniques in replicating the MC estimate while maintaining the functional and computational benefits of a direct method.
This enables robust, risk-aware, continuous motion-planning for a broad class of nonlinear and/or partially-observable systems.
\end{abstract}


\section{Introduction}

Robotic motion planning is a decision-making problem that must balance optimality and safety.
In the real world, these decisions are complicated by the presence of uncertainty due to imperfect sensing, partial observability, and stochastic dynamics.
This uncertainty is often difficult or impossible to explicitly bound, and safety cannot be guaranteed against all realizations of noise and disturbance.

This motivates the replacement of deterministic safety constraints with \emph{risk constraints} \cite{li2002probabilistically} that seek to compute or bound the probability of failure.
Unfortunately, exact evaluation of this probabilistic risk is challenging and computationally intractable for generic nonlinear systems.
While Monte-Carlo (MC) estimation techniques \cite{calafiore2006scenario,blackmore2010probabilistic,janson2018monte} provide a general and powerful workaround, they are still computationally-demanding and difficult to embed within a continuous motion planner.
A number of ``directly-computable'' risk approximations have been proposed \cite{li2002probabilistically,blackmore2009convex,patil2012estimating,ono2015chance}, but all fundamentally require that the system evolves in discrete time.
Many systems we care to control in practice evolve continuously, and while application of discrete-time methods is possible in these settings, the ensuing risk estimates will be highly sensitive to the chosen time discretization.
As recognized by \citet{ariu2017chance}, \citet{janson2018monte}, and others, they may be either too lax (allowing ``corner-cutting''), too conservative (leading to severe sub-optimality or artificial infeasibility), or both simultaneously.

\begin{figure}[t]
  \centering
  \includegraphics[width=0.9\linewidth]{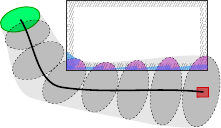}
  \caption{Consider a planar system maneuvering from the green initial distribution to the red goal position in the presence of a rectangular obstacle.
           Shaded ellipses represent the state distribution at a set of discrete timesteps, $\pT = \{t_0, t_1, \ldots, t_k\}$, and reflect uncertainty under closed-loop execution.
           A common technique for computing the total collision probability leverages Boole's inequality to simply sum the violation probabilities at each step (magenta).
           However, the resulting estimate will be sensitive to the choice of $\pT$ -- too coarse, and it will underestimate; too fine, and it will ``double-count'' probability mass corresponding to trajectories that remain in collision across multiple timesteps.
           This leads to over-conservatism, artificial infeasibility, and ultimately brittle planning.
           In this paper we introduce a \emph{continuous}-time approximation (cyan) that instead aggregates probability mass \emph{entering} collision over each \emph{interval}.
           As $\pT$ is refined, our result actually improves, eventually converging to the true failure probability.
          }
  \label{fig:intro_comparison}
\end{figure}

This paper addresses this problem at its source, taking a rigorous look at the evolution of failure probability directly in continuous time.
Our approximation is general, applying to partially-observable\footnote{
          Note that under partial-observability, the process $\x(t)$ can only be considered Markov if it is ``augmented'' with the internal state of the estimator/controller.
          For the purposes of this paper we leave this implicit -- because constraints are assumed to involve only the state of the physical plant, augmentation will contribute no analytic or computational cost.
  } stochastic systems $\x(t)$ in $\manX = \R^{n_x}$ with nonlinear It\^o dynamics
\begin{align}
  \diff \x(t) &= \vc{f}\big(\x(t), \u(t)\big) \diff t + \mx{G}\big(\x(t), \u(t) \big) \diff \w(t)  \label{eq:dynamics}
\end{align}
where $\u(t)$ and $\w(t)$ represent a vector-valued control process and Brownian motion, respectively.
Under partial observability, $\x(t)$ is not directly available to the controller, and feedback must be accomplished via a parallel observation process represented by the filtration $\Ft$ (the information available at time $t$).
The special case of fully-observable systems is easily captured here as well.

Our main results impose some additional restrictions between the dynamics (\ref{eq:dynamics}) and the constraints that define the \emph{safe set} $\Xsafe \subset \manX$.
Let $\Xsafe$ be defined as the intersection of sub-level sets $\{g_j(\x) \leq 0\}$ for $j = 1, \ldots, m$, and assume each $g_j : \manX \mapsto \R$ is at least twice-differentiable.
Crucially, we will require that the ``constraint space'' be separated from the the effect of the control $\u(t)$ and the stochastic disturbance $\w(t)$ by at least two derivatives --
\begin{assumption}\label{ass:second_order}
  Specifically, we will require that for each constraint $g_j(\x)$ with gradient $\vc{a}_j(\x)$
  \begin{equation}\label{eq:local_determinism}
    \vc{a}_j(\x)^\T \mx{G}(\x, \u) = 0 \quad \forall \x,
  \end{equation}
  and the constraint Lie derivative
  \begin{equation}\label{eq:hj_def}
    h_j(\x) \triangleq \rm{L}_{\vc{f}} g_j(\x) = \vc{a}_j(\x)^\T \vc{f}(\x, \u)
  \end{equation}
  is independent of the control $\u$ input.
\end{assumption}

Let $y_j(t) = g_j\big( \x(t) \big)$ represent the projection of $\x(t)$ in the output space of $g_j$, and let $\y(t)$ represent the stacked vector of such projections.
We will refer to $\y(t)$ as the constraint-space process, a natural object of consideration as $\x \in \Xsafe \iff \y \in \setOm$, where $\setOm \subset \R^m$ is the (closed) non-positive orthant.
Assumption \ref{ass:second_order} can be understand to require that $\y(t)$ evolves according to $\dot{\y}(t) = \h\big( \x(t) \big)$, and only depends on $\u(t)$ and $\w(t)$ indirectly through $\x(t)$.
This separation has the effect of modulating the contribution of the Brownian noise $\w(t)$, and provides a critical regularity condition to enabling the results in this paper.
In practice, this assumption reflects to the ubiquitous setting of physical systems with position constraints (\eg, mobile robots), in which control inputs and stochastic disturbances act in the space of forces and torques.
Assumption \ref{ass:second_order} has the added benefit of mitigating any sensitivity of our results to the particulars of the input process $\u(t)$, allowing open-loop input trajectories and state- or output-feedback policies to be handled identically in the analysis.

We will require one more technical consideration with respect to the constraint-space process $\y(t)$.
\begin{assumption}\label{ass:h_lipschitz}
  With respect to the constraint set $\{g_j(\x)\}$ and $\vc{f}(\x, \u)$, the constraint-space time-derivative $\h(\x)$ defined element-wise in (\ref{eq:hj_def}) is Lipschitz-continuous in $\x$.
  Specifically, there exists an $L > 0$ such that
  \begin{equation*}
    \norm{\h(\x_1) - \h(\x_2)} \leq L \norm{\x_1 - \x_2} \quad \forall \x_1, \x_2 \in \manX.
  \end{equation*}
\end{assumption}

The types of planning problems we address take the form
\begin{align}
  \min_{\u(\cdot)} \qquad &\E \int_0^T l\big(t, \u(t), \x(t) \big) \diff t  \label{eq:nom_cost} \\
  \text{subject to: }  & P\bigg( \bigvee_{t \in [0, T]} \x(t) \not \in \Xsafe \bigg) \leq \Delta  \label{eq:ct_constraint}
\end{align}
where the $\bigvee$ symbol is a logical OR, implying existence of a satisfying event among a (possibly uncountable) collection, and $\Delta \in (0,1)$ represents a given risk tolerance over the finite horizon $[0,T]$.
The constraint (\ref{eq:ct_constraint}) upper-bounds the total probability of failure at any instant in the planning horizon.
This probabilistic \emph{risk} is challenging to evaluate and optimize against because it couples states across the planning horizon as a whole.
As pointed out by \citet{ono2015chance} and others, joint constraints of this nature can be approached via Lagrangian relaxation; that is, converting the risk-\emph{constrained} problem to a risk-\emph{minimizing} one with objective
\begin{equation}\label{eq:lagrangian_opt}
  \E \int_0^T \!\!\!l\big(t, \u(t), \x(t)\big) \diff t + \lambda \Big[ P\big( \!\!\!\bigvee_{t \in [0, T]} \!\!\!\!\x(t) \not \in \Xsafe \big) - \Delta \Big]
\end{equation}
for some $\lambda \geq 0$.
However, the augmented objective in (\ref{eq:lagrangian_opt}) does not possess the time-additive \emph{Bellman} structure of (\ref{eq:nom_cost}), precluding the application of many optimal control techniques such as dynamic programming \cite{todorov2005generalized,van2012motion,ono2015chance}.
Like \cite{blackmore2009convex,ono2015chance} and other direct approximations, the approximation presented in this paper restores this convenient time-additive structure, but does so in a way that preserves applicability to continuous-time systems.

This paper takes a fresh look at the time-evolution of the failure probability in (\ref{eq:ct_constraint}), specifically in continuous-time.
Related work is outlined in Section \ref{s:related_work}.
Section \ref{s:time_additive} leverages the language of first-exit times to produce a time-additive framework for continuous-time risk estimation.
The classic theory is rich and well-explored, but to the best of the authors' knowledge no techniques yet exist to enable computation in the context of generic nonlinear systems and nonlinear constraints.
To address this, we propose a piecewise-continuous approximation in Sections \ref{s:piecewise} and \ref{s:computation} that provably converges as the time discretization is refined -- this allows us to losslessly extend classic results for constant-coefficient systems.
As a second contribution, we identify a lightweight method to account for ``safety-thus-far'' that avoids attempting to approximate an explicit posterior and ensures conservatism without being excessively so.
Finally, the resulting risk approximation is empirically demonstrated in Section \ref{s:experiments} to well-approximate MC estimates while retaining the computational simplicity and general applicability of direct methods, paving the way for risk-aware, continuous motion planning onboard real-world systems.
For the interested reader, \refAppRedQuad{} shows how the requisite numerical quadratures can be computed at significantly-reduced dimension, which is critical to ensure computational feasibility.

\section{Related Work}\label{s:related_work}
  Risk-aware motion planning is far from a new problem and has received much attention over the years.
  For example, when probabilistic constraints $P\big( \x(t) \not \in \Xsafe \big) \leq \Delta$ are enforced \emph{independently} for each $t$, suitable extensions of classic algorithms such as RRT \cite{luders2010chance,bry2011rapidly,van2011lqg} and differential dynamic programming (DDP) \cite{van2012motion} have been proposed.
  Alternatively, for problems with discrete time and action spaces, risk-constrained search methods such as RAO$^\star$ \cite{santana2016rao, huang2018hybrid} can be applied.


  This paper considers problems continuous in time and action, where safety is most naturally expressed by the joint constraint (\ref{eq:ct_constraint}).
  The field of robust control provides some solutions against \emph{worst-case} (\ie, bounded) disturbances, for example \citet{majumdar2013robust} and \citet{lopez2019dynamic}.
  However, these approaches are generally restricted to fully-observable systems.
  Furthermore, their performance may be severely conservative in the average case, motivating a quantification of \emph{probabilistic} risk to allow explicit trade-off between safety and expedience during planning.

  \subsection{Sampling-Based Methods}
    MC techniques provide a general and powerful method for estimating failure probabilities, at the cost of having to run a potentially large number of simulation rollouts.
    As closed-loop partially-observable systems require simulation of plant, estimator, and controller, this can represent non-negligible amounts of computation.
    Aside from the issue of sample-complexity, which has been partially addressed by \citet{calafiore2006scenario} and \citet{janson2018monte}, sample-based estimates are discontinuous and therefore fundamentally cumbersome to incorporate into a continuous optimization framework.
    For example, \cite{calafiore2006scenario} proposes enforcement of a deterministic constraint for each sample, \cite{janson2018monte} resorts to iterative obstacle inflation in RRT, and \citet{blackmore2010probabilistic} apply Mixed-Integer techniques.
    This limitation of MC motivates the search for efficient, ``direct'' risk approximations amenable to online, \emph{continuous} optimization over motion plans.

  \subsection{Direct Risk Estimates (in Discrete Time)}
    A number of direct techniques have been developed in a discrete-time setting, where (\ref{eq:ct_constraint}) simplifies to
    \begin{equation}\label{eq:dt_constraint}
      P\big( \vee_{k = 0}^K \x(t_k) \not \in \Xsafe \big) \leq \Delta.
    \end{equation}

    Early work by \citet{li2002probabilistically} recognized that, under Linear-Time-Varying (LTV) dynamics, dispersions will be distributed as an $n_x (K+1)$-dimensioned normal distribution.
    The joint probability in (\ref{eq:dt_constraint}) can then evaluated as a high-dimensional integral via quadrature methods or sampling, both of which have complexity exponential in dimension.
    To avoid this unfavorable scaling with $K$, \citet{blackmore2009convex} use Boole's inequality (a.k.a.\ the union bound) to decompose the probability in (\ref{eq:dt_constraint}) over time as
    \begin{equation}\label{eq:dt_booles}
      P\big( \vee_{k = 0}^K \x(t_k) \not \in \Xsafe \big) \leq \sum_{k = 0}^K P( \x(t_k) \not \in \Xsafe ).
    \end{equation}
    This decoupling over timesteps is highly convenient.
    As pointed out by \citet{ono2015chance}, the right-hand side of (\ref{eq:dt_booles}) is time-additive, allowing the use of dynamic programming (\ie, the Bellman principle) to minimize (\ref{eq:lagrangian_opt}).
    Alternatively, \citet{ono2008iterative} introduce the \emph{risk allocation} formulation
    \begin{equation}\label{eq:dt_risk_allocation}
      P( \x(t_k) \not \in \Xsafe ) \leq \Delta_k \,\, \forall k \leq K \quad \text{and} \quad \sum_{k=0}^{K} \Delta_k = \Delta
    \end{equation}
    which explicitly allocates a risk ``budget'' between timesteps.
    This framework has inspired a series of other works including \cite{vitus2011feedback,ma2012fast,dai2019pcheckov}, all fundamentally dependent on decomposition (\ref{eq:dt_booles}).

    Despite its popularity, the use of Boole's inequality to decompose joint constraints over time can be severely conservative, as illustrated in Fig.~\ref{fig:intro_comparison}.
    By ignoring correlations between states at adjacent timesteps, the sum in (\ref{eq:dt_booles}) will ``double-count'' failure events corresponding to violation in multiple (possibly adjacent) time intervals, particularly as the time discretization is refined.
    This problem is addressed by \citet{patil2012estimating}, who recognize that the state distributions should be \emph{conditioned} on the safety of prior timesteps.
    Because capturing this conditioning exactly is challenging, they approximate the projected and truncated distribution corresponding to each constraint as a single-dimensional Gaussian, allowing closed-form update of the state distribution parameters.
    However, Gaussianity here is inexact and the resulting estimate may not remain statistically consistent or result in an upper-bounding risk estimate.

  \subsection{Stochastic Processes in Continuous-Time}
    When addressing continuous-time systems, a standard approach approximates the original system as discrete-time under some discretization $\pT = \{t_0 = 0, t_1, \ldots, t_K = T\}$ and then substitutes constraint (\ref{eq:dt_constraint}) for (\ref{eq:ct_constraint}).
    This allows well-studied discrete-time techniques to be applied, but, as mentioned earlier, the results will be highly sensitive to the choice of discretization.
    Depending on the complexity of the environment and the movement speed of the robot, these effects can lead to unsafe or overly-cautious behavior, or both.

    In an effort to explicitly address the continuous-time constraint (\ref{eq:ct_constraint}) and avoid ``corner-cutting,'' \citet{ariu2017chance} apply Boole's inequality over \emph{intervals} rather than instantaneous states as
    \begin{equation}\label{eq:interval_booles}
      P\Big( \!\!\! \bigvee_{s \in [0,T]} \!\!\! \x(s) \not \in \Xsafe \Big) \leq \sum_{k=0}^{K-1} P\Big( \!\!\! \bigvee_{s \in [t_k,t_{k+1})} \!\!\!\!\! \x(s) \not \in \Xsafe \Big).
    \end{equation}
    Note that continuity allows the endpoint $\x(T)$ to be dropped.
    As with (\ref{eq:dt_booles}), this ``interval Boole's'' neglects correlations between events.
    In particular, their method is required to assume that the state process itself follows a Brownian motion, and ultimately results in a \emph{doubly}-conservative estimate compared to its discrete-time counterpart.
    By comparison, the approximation proposed in this paper also operates on intervals, but it addresses the nonlinear dynamics directly and avoids the use of Boole's inequality in this way, producing a much tighter risk bound.

    This paper leans heavily on the classical notion of first-passage times, a well-studied topic in the field of continuous stochastic processes.
    Indeed, the piecewise approximation presented in Section \ref{s:piecewise} is similar in spirit to that of \citet{jin2017first}, although we consider the case where the process (rather than purely the boundary) is nonlinear.

\section{How Risk Evolves in Time}\label{s:time_additive}
  Consider a \cadlag\footnote{a.k.a.\ right-continuous with left limits.}, Markov process $\z(t) \in \R^m$ with associated probability space $(\Omega, \mc{F}, P)$.
  Here, $\Omega$ represents the \emph{outcome} space, $\mc{F}$ is a \emph{sigma-algebra} over $\Omega$ (a collection of events $E \subseteq \Omega$) such that $\z(t)$ is measurable for all $t$, and $P$ is a probability measure over $\mc{F}$.
  Note that for now we do \emph{not} assume $\z(t)$ has continuous sample paths.
  We begin by exploring the ``nearly'' time-additive evolution of the \emph{exit cumulant} with respect to a closed set $D \subset \R^m$.
  \begin{equation}
    \Fz[\z](t) \triangleq P\big(\bigvee_{s \in [0,t]} \z(s) \not \in D \big).
  \end{equation}

  For clarity in the following discussion, we adopt the language of \emph{passage times}, also known as \emph{hitting} or \emph{exit times} \cite{karatzsas1988,yong1999}.
  Define a parameterized family of exit times with respect to process $\z(t)$ as
  \begin{equation}
    \tc[\z]_t \triangleq \inf \big\{s \in [t, T] \,\big|\, \z(s) \not \in D\big\}
  \end{equation}
  where the infinum of the empty set is assigned to $\infty$.
  That is, $\tc[\z]_t$ refers to the first instant (at or after $t$) that $\z(s)$ ``exits'' $D$.
  For both $\Fz[\z]$ and $\tc[\z]_t$ we will drop the explicit $\z$ specification when the corresponding process is unambiguous.

  The following useful properties can be identified.
  \begin{lemma}\label{lemma:tc_props}
    For any $0 \leq t_1 < t_2 \leq T$, the following hold:
    \begin{itemize}
      \item $\tc_{t_1} \geq t_2 \iff \z(s) \in D \quad \forall s \in [t_1, t_2)$.
      \item $\tc_0 \in [t_1, t_2) \iff \tc_0 \geq t_1, \tc_{t_1} < t_2$.
      \item For any outcome $\omega \in \Omega$, either
        \begin{enumerate}
          \item $\tc_0 = \infty$ (``no exit''), or
          \item $\z(\tc_0) \in \partial D \cup D^c$ (``smooth exit'' or a ``jump out'').
        \end{enumerate}
        where $\partial D$ and $D^c$ are the boundary and complement of $D$, respectively.
    \end{itemize}
  \end{lemma}

  It is straightforward to verify that the total exit probability can be written as $\Fz(T) = P(\tc_0 \leq T)$, and thus $\tc_0$ offers a means by which to analyze the time-evolution of $\Fz(t)$.
  In some cases $\Fz(t)$ is known to be time-differentiable \cite{jin2017first}, and thus computing this derivative (called the \emph{first-passage density}) would seem to be a natural goal.
  However, this computation is challenging for generic nonlinear processes, and instead we will settle for an interval-based ``integration'' scheme that reflects how $\Fz(T)$ can be approximated in practice.
  In later sections, through both analysis and experiment we demonstrate that this integrated approximation indeed converges to the true $\Fz(T)$ as the time discretization is refined.

  Proceeding, assume a given partition $\pT = \{t_0 = 0, t_1, t_2, \ldots, t_K = T\}$ of the fixed horizon $[0,T]$.
  A crucial advantage of the language of the \emph{first}-exit time is that it provides a natural disjointness between events, and in particular
  \begin{align}
    \Fz(T) 
           &= \sum_{k=0}^{K-1} P\big(\tc_0 \in [t_k, t_{k+1})\big) + P(\tc_0 = T)  \label{eq:ival_cond1}.
  \end{align}
  Note that, in contrast to (\ref{eq:interval_booles}), the relation (\ref{eq:ival_cond1}) holds with equality.
  Proceeding from here and adopting the shorthand $\z_k \triangleq \z(t_k)$, $\tc_k \triangleq \tc_{t_k}$, and so on, the above can be written
  \begin{align}
    &\sum_{k=0}^{K-1} \Big( P\big(\tc_0 \in [t_k, t_{k+1}), \z_k \in D\big) + P(\tc_0 = t_k, \z_k \in D^c) \Big) \nonumber \\
      &\qquad \qquad  + P(\tc_0 = T, \z(T) \in D^c)  \label{eq:pt:marginalize_endpoint}  \\
    &= \sum_{k=0}^{K-1} P\big(\tc_0 \in [t_k, t_{k+1}), \z_k \in D \big) + \sum_{k=0}^K P(\tc_0 = t_k, \z_k \in D^c)  \label{eq:Fz_sum0}
  \end{align}
  where in (\ref{eq:pt:marginalize_endpoint}) we split the probability over the event that $\z_k \in D$.
  Note that the right-hand summation in (\ref{eq:Fz_sum0}) involves the probabilities that discontinuous sample paths ``jump out'' of $D$ at any of the partition points $t_k$.
  Examining the first set of terms, Lemma \ref{lemma:tc_props} allows
  \begin{align}
    &P\big(\tc_0 \in [t_k, t_{k+1}), \z_k \in D \big)    \label{eq:Phi_deriv} \\
    &= P\big(\tc_k < t_{k+1} \cbar \tc_0 \geq t_k, \z_k \in D \big) P(\tc_0 \geq t_k, \z_k \in D). \nonumber
  \end{align}
  The conditioning on $\tc_0 \geq t_k$ and $\z_k \in D$ in (\ref{eq:Phi_deriv}) implies that the process has not exited ``yet,'' imposing a specific posterior over $\z_k$ that we will call the \emph{anthropic distribution}\footnote{
    In cosmology, the \emph{anthropic principle} remarks that life can only observe universes that themselves allow for the existence of life.
  }
  \begin{equation}\label{eq:anthropic_dist}
    \pi_t(\diff \z) \triangleq P(\z(t) \in \diff \z \cbar \tc_0 \geq t, \z(t) \in D).
  \end{equation}

  Before proceeding further, define the function
  \begin{equation}\label{eq:Phiz_def}
    \Phiz(t_k, t_{k+1}; \mu_k) \triangleq \int_{\z} P\big(\tc_k < t_{k+1} \cbar \z_k = \z\big) \diff \mu_k( \z )
  \end{equation}
  to represent the probability of an exit (not necessarily the first) in the one-sided interval $[t_k, t_{k+1})$, given that $\z_k$ is distributed according to the distribution $\mu_k$ over $\R^m$.
  Note that the distribution $\mu_k$ need not represent a normalized probability distribution.
  Because the process $\z(t)$ is assumed Markov and applying (\ref{eq:Phi_deriv}), sum (\ref{eq:Fz_sum0}) can be re-written in terms of $\Phiz$ and $\pi_t$ as
  \begin{align}
    \Fz(T) &= \sum_{k=0}^{K-1} \Phiz(t_k, t_{k+1}; \pi_k) P( \tc_0 \geq t_k, \z_k \in D) \notag \\
           &~~+ \sum_{k=0}^K P(\tc_0 = t_k, \z_k \in D^c). \label{eq:Fz_sum2}
  \end{align}


  A key challenge in evaluation of (\ref{eq:Fz_sum2}) is computation of $\pi_k$.
  For one thing, its support is clearly limited to $D$, which is sufficient to ensure non-Gaussianity.
  This motivates us to avoid attempting to approximate or bound $\pi_k$ directly and instead decompose it via Baye's rule, producing
  \begin{align}
    \pi_k( \diff \z ) 
                  &= \frac{ P\big( \tc_0 \geq t_k, \z_k \in D \cbar \z_k = \z \big) P\big(\z_k \in \diff \z \big) }{P( \tc_0 \geq t_k, \z_k \in D )}  \nonumber \\
                  &\triangleq \frac{ \Psi_k(\z) b_k(\diff \z) }{ P(\tc_0 \geq t_k, \z_k \in D). }  \label{eq:pi_decomp}
  \end{align}
  where $b_k$ is the \emph{a priori} distribution of $\z_k$, and $\Psi_k : \R^m \mapsto [0, 1]$ will be referred to as the \emph{anthropic likelihood}.
  It is straightforward to see
  \begin{align}
    &\Phiz(t_k, t_{k+1}; \pi_k) P( \tc_0 \geq t_k, \z_k \in D )   \nonumber \\
    &\quad = \int_{\z} P(\tau_k \leq t_{k+1} \cbar \z_k = \z) P( \tc_0 \geq t_k, \z_k \in D ) \diff \pi_k(\z)   \nonumber \\
    &\quad = \int_{\z} P(\tau_k \leq t_{k+1} \cbar \z_k = \z) \diff \big( \Psi_k(\z) b_k(\z) \big)  \nonumber \\
    &\quad = \Phiz(t_k, t_{k+1}; \Psi_k b_k)  \label{eq:Psi_identity}
  \end{align}
  and therefore (\ref{eq:Fz_sum2}) can be written
  \begin{equation}\label{eq:Fz_sum}
    \Fz(T) = \sum_{k=0}^{K-1} \Phiz(t_k, t_{k+1}; \Psi_k b_k) + \sum_{k=0}^K P(\tc_0 = t_k, \z_k \in D^c).
  \end{equation}
  As discussed in Section \ref{ss:anthro_approx}, and in contrast to the case of $\pi_k$, identifying conservative approximations for $\Psi_k$ will be both straightforward and effective.

  \subsection{Analogous Development Under Runtime Information}
    Though not the primary focus of this paper, we note that the preceding development can be applied analogously in the context on a filtration $\Ft$ representing information that becomes available \emph{during} execution as opposed to \emph{a priori}, for example from sensor observations.

    Though the information generated by $\Ft$ itself evolves randomly, modification of the above discussion is straightforward and we simply provide some analogous definitions here for clarity.
    As in (\ref{eq:anthropic_dist}), we can define the \emph{anthropic belief}
    $  \bar{\pi}_t(\diff \z) \triangleq P\big( \z(t) \in \diff \z \cbar \tc_0 \geq t, \z(t) \in D, \Ft \big)$,
    and like (\ref{eq:anthropic_dist}) it has the structure
    \begin{equation}\label{eq:anthro_belief}
      \bar{\pi}_t(\diff \z) = \frac{\bar{\Psi}_t(\z) \bar{b}_t(\diff \z)}{ P\big(\tc_0 \geq t, \z(t) \in D \cbar \Ft \big) }
    \end{equation}
    where the estimated \emph{belief} $\bar{b}_t$ and anthropic likelihood $\bar{\Psi}_t$ are adapted to $\Ft$.
    An analogous identity to (\ref{eq:Psi_identity}) can be established, yielding
    \begin{equation}\label{eq:Fz_sum_partial_obs}
      \Fz(T) = \E \Big[ \sum_{k = 0}^{K-1} \Phiz(t_k, t_{k+1}; \bar{\Psi}_k \bar{b}_k) \Big] + \sum_{k=0}^K P(\tc_0 = t_k, \z_k \in D^c).
    \end{equation}
    where the expectation is taken over $\Ft$.
    Crucially, (\ref{eq:Fz_sum_partial_obs}) possesses the same Bellman structure as the fully-observable case (\ref{eq:Fz_sum}).


\section{A Piecewise-Continuous Approximation}\label{s:piecewise}
  The preceding section introduced a framework for decomposing the failure probability for a generic process $\z(t)$ according to a discrete time partition $\pT$.
  Rather than attempting to apply the decomposition (\ref{eq:Fz_sum}) directly in the space of the nonlinear process $\x(t)$, this section introduces an approximation of the constraint-space process $\y(t)$ for which computation can be made tractable.

  Assume a given time discretization $\pT = \{t_0 = 0, \ldots, t_k, t_{k+1}, \ldots, t_K = T\}$.
  Define an approximating process $\yhat(t)$ such that, for a given $s \in [t_k, t_{k+1})$,
  \begin{equation}
    \yhat(s) = \y(t_k) + (s-t_k) \vc{h}(\x_k)
  \end{equation}
  where the vector $\h(\x)$ is defined element-wise in (\ref{eq:hj_def}).
  This simplified process is piecewise-linear according to the given time discretization $\pT$, which controls the ``accuracy'' of the approximation.
  Though $\pT$ in practice will likely be dictated by the computational resources available, it will be shown that as it is refined the corresponding failure probability estimate $\Fz[\yhat](T)$ converges to the true failure probability $\Fz[\y](T) = \Fz[\x](T)$.
  That is to say, this approximation is asymptotically lossless.

\begin{figure}[t]
    \centering
    \begin{tikzpicture}[scale=1, every node/.style={transform shape}]
      \node[anchor=south west,inner sep=0] (image) at (0,0)
          {\includegraphics[width=\linewidth]{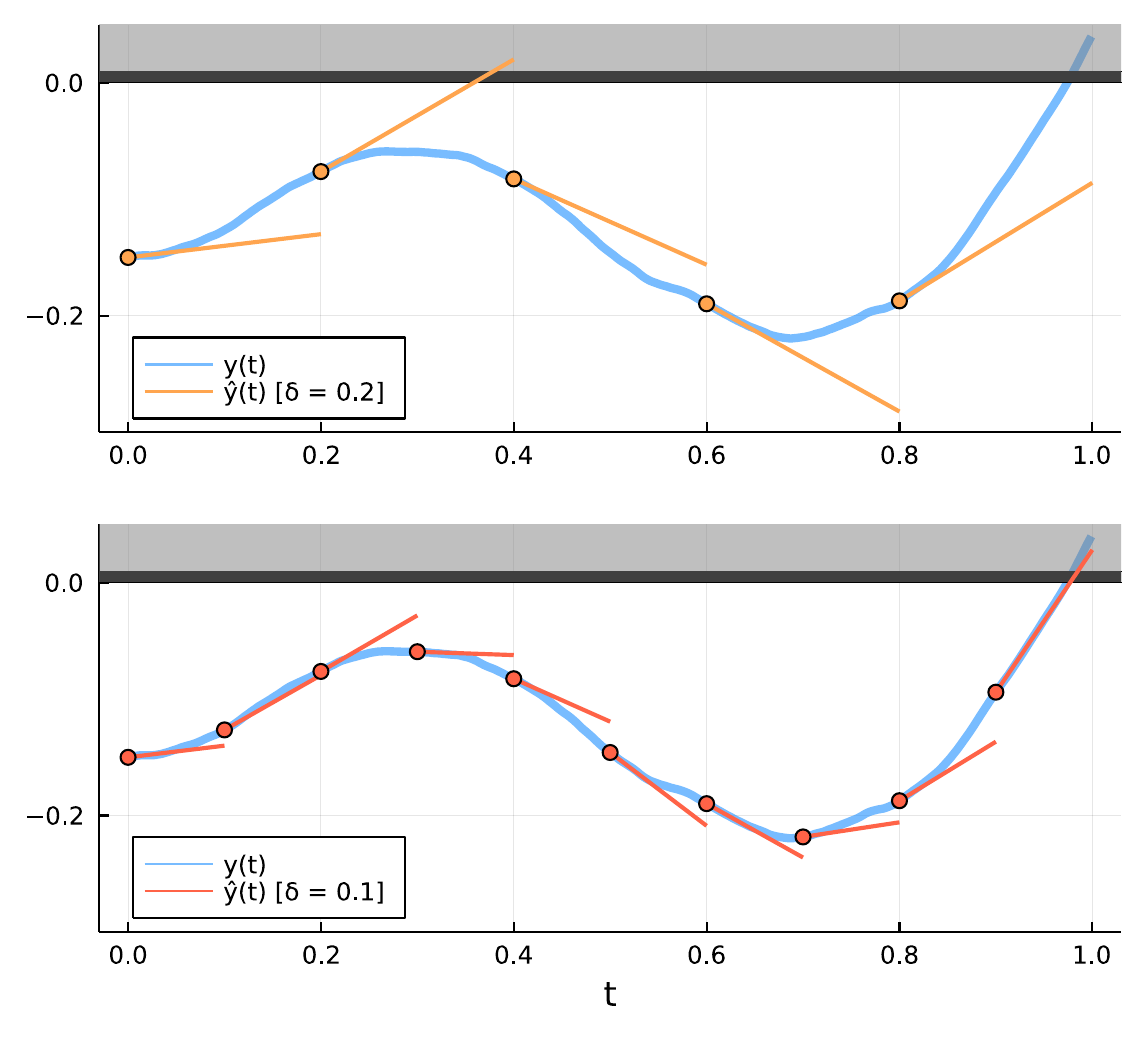}};
    \end{tikzpicture}
    \vspace{-0.9cm}
    \caption{
        Illustration of a random rollout of a one-dimensional process $y(t)$ and the corresponding piecewise-linear approximations $\hat{y}(t)$ for varying time discretizations $\pT$.
        The upper plot shows the approximation for a relatively coarse mesh (larger $\delta$), and the bottom for a relatively fine mesh (smaller $\delta$).
        In either case, the original and approximating processes coincide at each timestep $t_k \in \pT$, and as $\pT$ is refined (that is, as $\delta \rightarrow 0$), the approximation $\hat{y}(t)$ converges to the true process $y(\cdot)$.
        Upward zero-crossings of $y(t)$ imply constraint violations, and our results show that the true failure probability (that of process $y(t)$) can be well-approximated via the much simpler process $\hat{y}(t)$.
    }
    \label{fig:yhat}
  \end{figure}

  Fig.~\ref{fig:yhat} illustrates how the approximation $\yhat(t)$ follows $\y(t)$ for a given $\pT$, and how this approximation improves as $\pT$ is refined.

  \subsection{Convergence Analysis}
    Consider evaluating $\Fz[\yhat](T)$ for the approximating process $\yhat(t)$ -- the result will depend on the underlying time discretization $\pT$ used to construct $\yhat$.
    In this section, we show that as $\pT$ is refined (that is, as mesh size $\delta \rightarrow 0$), $\yhat(t)$ converges to $\y(t)$ in a specific sense, and furthermore $\Fz[\yhat](T)$ converges to $\Fz[\y](T)$.
    Proofs are left to \refAppProofs.

    \begin{definition}[Pathwise-Uniform Convergence]\label{defn:puc}
      For a sequence of processes $\{\zn(t)\}$, we say $\zn(t)$ converges \emph{pathwise-uniformly} to process $\z(t)$ if
      there exist some real-valued, decaying sequence $\epsn \rightarrow 0$ such that
      \begin{equation*}
        P \Big( \sup_{t \in [0,T]} \norm{\zn(t) - \z(t)} > \epsn \Big) \leq \epsn \rightarrow 0.
      \end{equation*}
    \end{definition}
    Pathwise-uniform convergence (PUC) requires that both the uniform error bound and the corresponding probability of excessive error go to zero simultaneously.
    Note that PUC does \emph{not} imply that $\zn(\cdot; \omega)$ converges to $\z(\cdot; \omega)$ for any fixed outcome $\omega \in \Omega$
      -- nevertheless, we will see that this condition is sufficient to ensure a non-trivial convergence-of-probabilities result.

    \begin{proposition}\label{prop:puc_implies_P_conv}
      If a given sequence of $\R^m$-valued processes $\zn(t)$ converges pathwise-uniformly to $\z(t)$, and under the mild regularity condition that $P\Big( \sup_{t \in [0,T]} z_j(t) = 0 \Big) = 0$ for all $j$, then
      \begin{equation*}
        P\Big(\bigvee_{t \in [0,T]} \zn(t) \not \in \setOm \Big) \rightarrow P \Big( \bigvee_{t \in [0,T]} \z(t) \not \in \setOm \Big).
      \end{equation*}
    \end{proposition}

    Thus, if we can show that the sequence of approximating constraint-space processes $\yhatn(t)$ produced by refining $\pT$ converges pathwise-uniformly to the true constraint-space process $\y(t)$, Proposition \ref{prop:puc_implies_P_conv} ensures that $\Fz[\yhat](T) \rightarrow \Fz[\y](T)$.
    This leads to our main results.
    \begin{proposition}\label{prop:yhatn_puc}
      Let Assumptions \ref{ass:second_order} and \ref{ass:h_lipschitz} hold with respect to the process $\x(t)$ and the $m$ constraint functions $\{g_j\}$.
      Additionally, assume the following mild regularity conditions:
      \begin{enumerate}
        \item $\E \big[ \vc{f}\big(\x(t), \u(t)\big) \big] \leq c$ for all $t \in [0,T]$,
        \item $\E \norm{ \mx{G}\big( \x(t), \u(t) \big) }^2 \leq c^2$ for all $t \in [0,T]$.
      \end{enumerate}
      Consider a sequence of discretizations $\pTn = \{0 = t_0, t_1, \ldots, t_{\Kn} = T\}$ of the compact interval $[0, T]$ with mesh sizes $\deltan \rightarrow 0$ and segment count $\Kn \leq 2 T / \deltan$.
      Then the corresponding sequence of piecewise-linear approximating processes $\yhatn(t)$ converges pathwise-uniformly to the true constraint-space process $\y(t)$.
    \end{proposition}

    \begin{corollary}\label{cor:Fy_convergence}
      If the assumptions of Prop.~\ref{prop:yhatn_puc} hold and $P\Big( \sup_{t\in [0,T]} y_j(T) = 0 \Big) = 0$ for all $j$
      then the corresponding approximating failure probabilities $\Fz[\yhatn](T)$ converge to the true failure probability $\Fz[\y](T)$.
    \end{corollary}

    The additional regularity condition of Corollary $\ref{cor:Fy_convergence}$ is somewhat difficult to verify;
    for the purposes of the paper we consider this, as well as the regularity conditions required by Prop.~\ref{prop:yhatn_puc}, to be mild.

    Corollary \ref{cor:Fy_convergence} makes clear that the substitution of $\yhat(t)$ for $\y(t)$ is (asymptotically) loss-less for the purposes of computing the risk probability $\Fz[\y](T)$.
    It is worth noting that this result does not anywhere assume that the dynamics or observation model are Gaussian or LTV, although practical computation of distributions $b_k$ may still require this assumption.

    There remains one lingering challenge however -- $\yhat(t)$ is only piecewise-continuous and thus the discontinuous ``jump out'' probabilities in (\ref{eq:Fz_sum})
    \begin{equation}\label{eq:jump_probs}
      \sum_{k=1}^K P(\tc_0 = t_k, \yhat_k \not \in \setOm)
    \end{equation}
    may be non-zero.
    These terms are difficult to compute in practice, as they essentially capture the effect of ``undetected'' zero-crossings of the true process $\y(t)$.
    However, the following conjecture suggests they can be safely ignored in the limit, and our empirical results confirm this in practice.
    \begin{conjecture}\label{conj:jump_probs_negligible}
      Let $N : \Omega \rightarrow \mathbb{Z}^+$ be a random variable indicating the number of exits of $\y(t)$ from the safe region $\setOm$.
      So long as $\E[ N ]$ is finite, then the jump probabilities sum (\ref{eq:jump_probs}) is bounded
      \begin{equation*}
        \sum_{k=1}^K P(\tc_0 = t_k, \yhat_k \not \in \setOm) \leq \E [ N ] < \infty,
      \end{equation*}
      and, in particular, the sum on the left decays to zero as $\delta \rightarrow 0$.
    \end{conjecture}
    A discussion and outline of a potential proof for Conjecture \ref{conj:jump_probs_negligible} is provided in \refAppProofs.

\section{Computing $\Fz[\yhat]$}\label{s:computation}
  The results of Section \ref{s:piecewise} indicate that we can replace the constraint-space process $\y(t)$ with the piecewise-linear $\yhat(t)$ for the purpose of computing failure probability, with the confidence that such approximation is asymptotically lossless.
  Furthermore, Conjecture \ref{conj:jump_probs_negligible} justifies dropping the ``jump out'' terms (\ref{eq:jump_probs}) arising from the discontinuity of $\yhat(t)$, leaving us to compute
  \begin{equation}\label{eq:Fyhat}
    \Fhat(T) \triangleq P(\x_0 \not \in \Xsafe) + \sum_{k = 0}^{K-1} \Phiz[\yhat](t_k, t_{k+1}; \Psi_k b_k).
  \end{equation}
  For simplicity, we will neglect the probability of $\x_0$ starting outside $\Xsafe$, allowing us to drop the first term throughout the rest of this paper.

  The piecewise-linear approximation $\yhat(t)$ makes computing $\Phiz[\yhat](t_k, t_{k+1}; \mu_k)$ straightforward from the definition (\ref{eq:Phiz_def})
  \begin{equation}\label{eq:Phiz_yhat}
    \Phiz[\yhat]\big(t_k, t_{k+1}; \mu_k\big) = \int_\x \Ind\big( \vc{g}(\x) + \delta_k \h(\x) \not \in \setOm \big) \diff \mu_k(\x)
  \end{equation}
  which can be computed via a numeric quadrature over the measure $\mu_k$, where $\delta_k = (t_{k+1} - t_k)$.
  In fact, the second-order nature of $\y(t)$ from Assumption \ref{ass:second_order} means that a relatively low-dimensional quadrature will suffice in practice
    -- more details are provided in \refAppRedQuad.

  Given the ability to compute (\ref{eq:Phiz_yhat}) from arbitrary measure $\mu_k$, all that remains is a practical means of approximating means of approximating our measure of interest: $\pi_k$ or $\Psi_k b_k$.

  \subsection{Approximating Anthropic Belief}\label{ss:anthro_approx}
    It is relatively straightforward to approximate the state distribution $b_k$ under the dynamics (\ref{eq:dynamics}), for example via a Gaussian distribution under a local LTV approximation.
    Nevertheless, an accurate estimate of total failure probability requires a means of capturing the survivorship bias and avoid double-counting -- this is the precisely what is captured in the anthropic state distribution $\pi_k$ or likelihood $\Psi_k$.
    As will be discussed, these functions are not well-captured via conventional Gaussian approximations, and a more special-purpose approximation will be identified.

    \subsubsection{Approximating $\pi_k$ as Gaussian}\label{sss:Gaussian_pi}
      In existing work, \citet{patil2012estimating} propose a Gaussian approximation for the anthropic distribution $\pi_k$ (or analogously, anthropic belief $\pi_k$) at each timestep $t_k$.
      This is attractive because it implies $\pi_k$ can be propagated in parallel with $b_k$.
      Specifically, consider approximating $\pi_k$ by conditioning on safety (via a recursive Gaussian approximation) at each prior timestep
      \begin{equation}\label{eq:pi_gauss}
        \hat{\pi}_k(\diff \x) \approx P\big(\x_k \in \diff \x \cbar \x_l \in \Xsafe \, \forall l \leq k \big).
      \end{equation}
      We refer the interested reader to their paper for implementation details.

      Again considering (\ref{eq:Fyhat}), we can apply identity (\ref{eq:Psi_identity}) and directly produce an upper-bound in terms of $\pi_k$
      \begin{equation}\label{eq:ival_gauss}
        \Fhat(T) \leq \sum_{k=0}^{K-1} \Phiz[\yhat](t_k, t_{k+1}; \pi_k).
      \end{equation}
      Together with our Gaussian approximation $\hat{\pi}_k$ from (\ref{eq:pi_gauss}), we refer to the resulting estimate as \texttt{ival\_gauss}.

      While the independence and Gaussianity assumptions involved here are convenient, they are also heuristic and non-conservative.
      This can lead to inconsistent $\pi_k$ estimates and ultimately unpredictable risk estimation, as will be demonstrated numerically in Section \ref{s:experiments}.

    \subsubsection{A Simple Alternative}\label{sss:ival_safe}
      Rather than attempting to represent and propagate $\pi_k$, we propose a simple, yet surprisingly powerful approximation of the anthropic likelihood $\Psi_k$.
      A clear upper-bound follows directly from the definition as
      \begin{equation}\label{eq:Psi_dumb}
        \Psi_k(\x) \triangleq P\big(\tc_0 \geq t_k, \x_k \in \Xsafe \cbar \x_k = \x \big) \leq \Ind_{\Xsafe}(\x)
      \end{equation}
      where $\Ind_{\Xsafe}$ is simply the indicator over the safe region $\Xsafe$.
      Applying (\ref{eq:Psi_dumb}) to (\ref{eq:Fyhat}) yields a conservative estimate
      \begin{equation}\label{eq:ival_safe}
        \Fhat(T) \leq \sum_{k = 0}^{K-1} \Phiz[\yhat](t_k, t_{k+1}; \Ind_{\Xsafe} b_k)
      \end{equation}
      which accumulates ``new'' exits over each interval -- note the contrast with the naive ``interval Boole's'' approximation given by (\ref{eq:interval_booles}).
      Because it restricts exit flow to probability mass which is ``safe'' at the start of each interval, we refer to approximation (\ref{eq:ival_safe}) as \texttt{ival\_safe}.

      Technically-speaking, (\ref{eq:ival_safe}) shares one of the same weaknesses of the discrete-time Boole's approximation in (\ref{eq:dt_booles}) -- it is not bounded above, and indeed may diverge as $\pT$ is refined.
      Nevertheless, it avoids the ``local'' double-counting of (\ref{eq:dt_booles}) by accumulating only probability mass \emph{leaving} $\Xsafe$ in each interval.
      This interpretation is illustrated in Fig.~\ref{fig:intro_comparison}, and our numeric results suggest \texttt{ival\_safe} does very well in practice.
      Furthermore, compared to \cite{patil2012estimating} or (\ref{eq:ival_gauss}) this approach requires no ``extra'' belief propagation at all, making it particularly lightweight.

  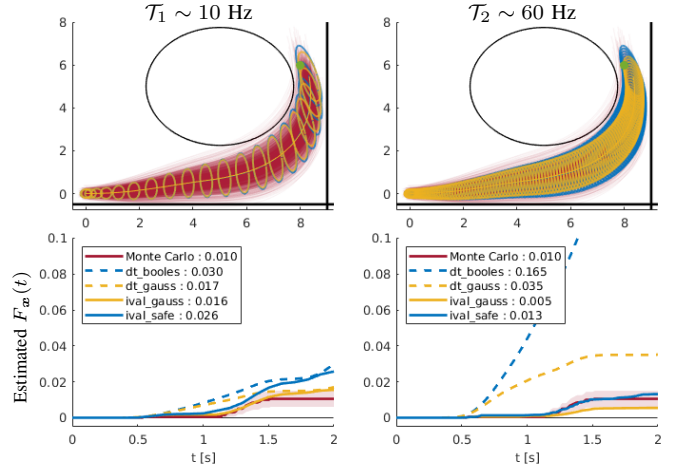
\begin{figure}[t]
    \centering
    \begin{tikzpicture}[scale=1.0, every node/.style={transform shape}]
      \node[anchor=south west,inner sep=0] (image) at (0,0)
          {\includegraphics[width=0.45\linewidth,trim={0.5cm, 0.4cm, 0.4cm, 0.37},clip]{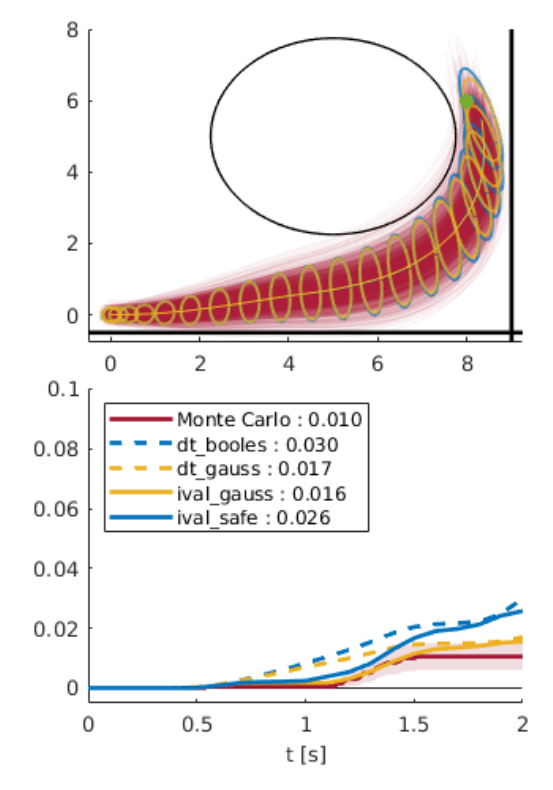}};
      \node at (2.15, 6) {\footnotesize $\pT_1 \sim 10$ Hz};
      \node[rotate=90] at (-0.2, 1.7) {\scriptsize Estimated $\Fz[\x](t)$};
    \end{tikzpicture}
    \hfill
    \begin{tikzpicture}[scale=1.0, every node/.style={transform shape}]
      \node[anchor=south west,inner sep=0] (image) at (0,0)
          {\includegraphics[width=0.45\linewidth,trim={0.5cm, 0.4cm, 0.4cm, 0.37},clip]{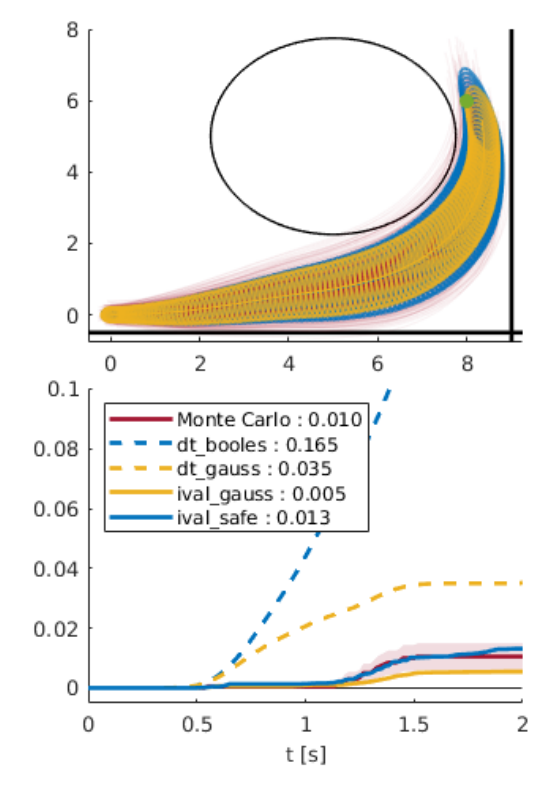}};
      \node at (2.15, 6) {\footnotesize $\pT_2 \sim 60$ Hz};
    \end{tikzpicture}
    \vspace{-0.25cm}
    \caption{
      We simulate a noisy second-order Dubin's car system navigating in a narrow passageway, computing collision probabilities under coarse (left) and fine (right) time discretizations, $\pT_1$ and $\pT_2$.
      MC trajectories (red) simulate a fixed 60Hz feedback control rate, and a corresponding \emph{a priori} state distribution $b_k$ (blue) is propagated via an LTV-Gaussian assumption.
      The anthropic estimates $\{\hat{\pi}_k\}_j$ (yellow) are computed according to each $\pT_j$.
      The underlying plots confirm the divergence of \texttt{dt\_booles} under fine discretizations, and illustrate the susceptibility of \texttt{dt\_gauss} and \texttt{ival\_gauss} to inconsistency in the Gaussian $\hat{\pi}_k$ estimates.
      In contrast, our method \texttt{ival\_safe} avoids estimating $\pi$ entirely, producing a lightweight and consistent risk estimate.
    }
    \label{fig:cc_demo}
  \end{figure}

\section{Numeric Results}\label{s:experiments}
  \begin{figure*}[t]
    \centering
    \begin{tikzpicture}[scale=1.0, every node/.style={transform shape}]
      \node[anchor=south west,inner sep=0] (image) at (0,0)
          {\includegraphics[width=0.225\linewidth,trim={0.5cm, 0.5cm, 0.5cm, 0.45cm},clip]{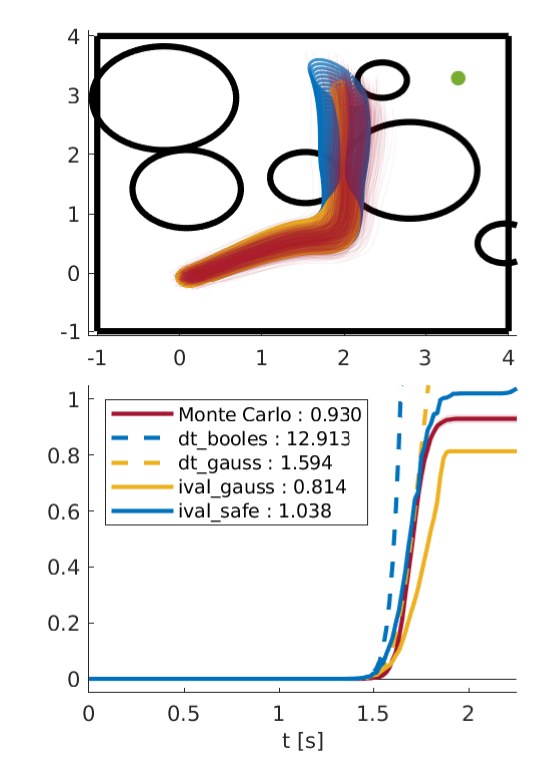}};
      \node[rotate=90] at (-0.1, 1.7) {\scriptsize Estimated $\Fz[\x](t)$};
    \end{tikzpicture}
    \begin{tikzpicture}[scale=1.0, every node/.style={transform shape}]
      \node[anchor=south west,inner sep=0] (image) at (0,0)
          {\includegraphics[width=0.225\linewidth,trim={0.5cm, 0.5cm, 0.5cm, 0.45cm},clip]{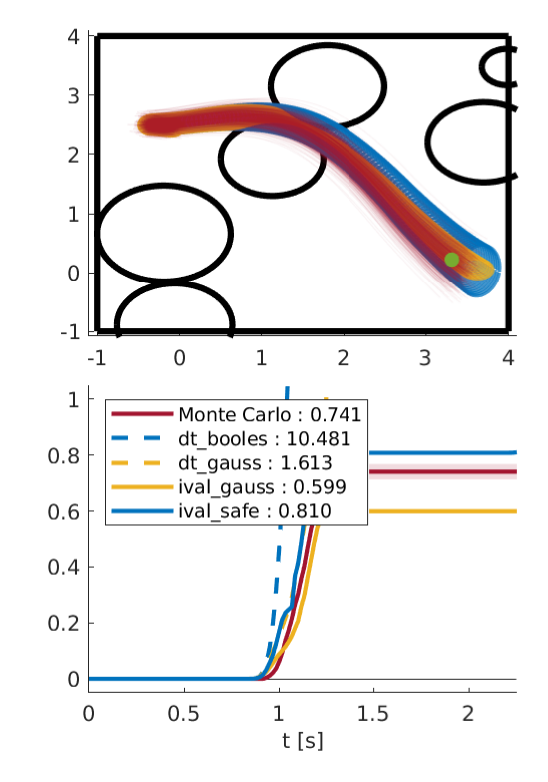}};
    \end{tikzpicture}
    %
    \hfill
    \vrule
    \hfill
    %
    \begin{tikzpicture}[scale=1.0, every node/.style={transform shape}]
      \node[anchor=south west,inner sep=0] (image) at (0,0)
          {\includegraphics[width=0.225\linewidth,trim={0.5cm, 0.5cm, 0.5cm, 0.45cm},clip]{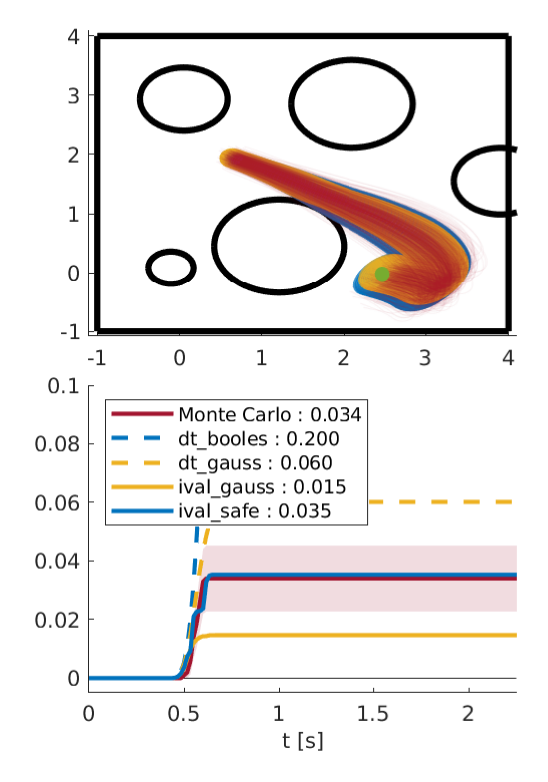}};
      \node[rotate=90] at (-0.2, 1.7) {\scriptsize Estimated $\Fz[\x](t)$};
    \end{tikzpicture}
    \begin{tikzpicture}[scale=1.0, every node/.style={transform shape}]
      \node[anchor=south west,inner sep=0] (image) at (0,0)
          {\includegraphics[width=0.225\linewidth,trim={0.5cm, 0.5cm, 0.5cm, 0.45cm},clip]{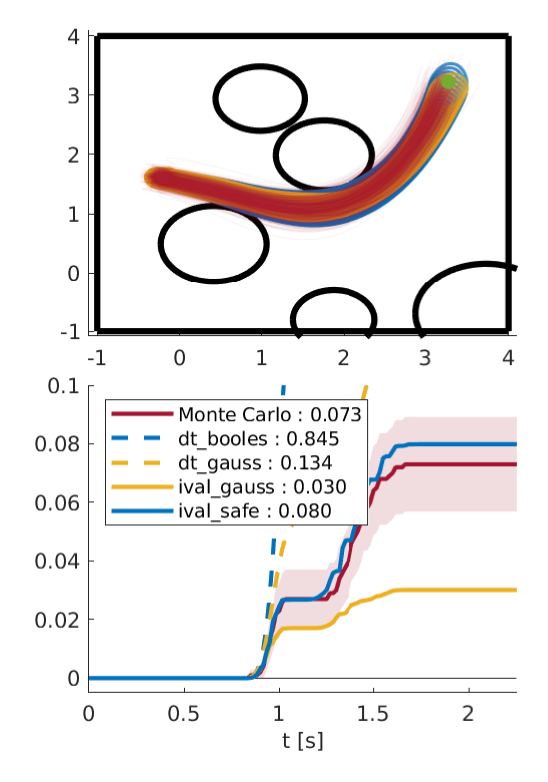}};
    \end{tikzpicture}
    \vspace{-0.2cm}
    \caption{
      Sample Dubin's car trajectories in randomly-generated environments over a $T = 2.5$ [s] horizon under 60 Hz LQG feedback control towards the (green) goal.
      The two scenarios on the left demonstrate \emph{nominally}-safe plans, which can experience significant risk upon stochastic execution.
      On the right are risk-constrained plans (generated using our \texttt{ival\_safe} approximation with $\Delta = 10\%$).
      MC trajectories are shown in red, and dispersions are well-approximated by the \emph{a priori} distributions $\{b_k\}$ shown in blue.
      The Gaussian anthropic belief estimates $\{\hat{\pi}_k\}$ are shown in yellow.
      The lower plots show $\Fz[\x](t)$ estimates as compared to MC (estimated standard error is indicated by the shaded region).
      Note that \texttt{dt\_booles} is prone to extreme over-estimation, while the two $\{\hat{\pi}_k\}$-dependent methods \texttt{dt\_gauss} and \texttt{ival\_gauss} suffer from the inconsistency of the underlying Gaussianity assumption.
      In contrast, our proposed method \texttt{ival\_safe} achieves the best approximation performance and avoids explicit $\pi_k$ approximation.
    }
    \label{fig:eval_samples}
  \end{figure*}

  \begin{table*}[h]
    \centering
    \caption{Compute times for Dubin's car MATLAB implementation on consumer laptop. QD refers to quadrature dimension.}
    \vspace{-0.3cm}
    \label{table:compute_times}
    \begin{tabular}{c  @{\quad\qquad\qquad}  c @{\qquad} c  @{\quad\qquad\qquad}  c @{\qquad} c}
      \toprule
      \vspace{0.07cm}
      Monte Carlo     &                          &                              &  $\sum_{k=0}^K P\big(x(t_k) \not \in \Xsafe\big)$  & $\sum_{k=0}^{K-1} \Phiz(t_k, t_{k+1}; \Ind_\Xsafe b_k)$  \\
      ($N = 1000$)    &  LTV-Gaussian $b_k$  &   Gaussian $\hat{\pi}_k$  &  QD: $(n_p-1) = 1$                            & QD: $n_p = 2$ \\
      \midrule
         39.1 [s]     &                0.034 [s] &                  0.054 [s]   &                                      0.034 [s]  &                   0.316 [s]   \\
      \bottomrule
    \end{tabular}
    \vspace{-0.3cm}
  \end{table*}

  To evaluate the accuracy of our approximation, we simulate a second-order Dubin's car system in the plane.
  System details are provided in \refAppSys.
  Noise in dynamics and observations render the system partially-observable, and an LQG-style feedback control policy stabilizes around a nominal trajectory.
  We evaluate four principle approximations.
  Discrete-time methods \texttt{dt\_booles} and \texttt{dt\_gauss} refer to the naive Boole's approximation (\ref{eq:dt_booles}) and the conditioned variant proposed by \cite{patil2012estimating}, respectively.
  Our methods \texttt{ival\_gauss} and \texttt{ival\_safe} are described in the preceding section.

  First, we use our \texttt{ival\_safe} estimate to optimize a risk-constrained ($\Delta = 2 \%$) plan in a tight environment, shown in Fig.~\ref{fig:cc_demo}.
  We then computed risk estimates for this plan under multiple time discretizations.
  As predicted, \texttt{dt\_booles} diverges severely as $\pT$ is refined while \texttt{ival\_safe} converges correctly.
  Interestingly, \texttt{dt\_gauss} and \texttt{ival\_gauss} converge to incorrect values, likely due to inconsistency in the underlying Gaussian $\hat{\pi}_t$ estimate.

  Next, we perform a larger statistical evaluation over random, programmatically-generated environments.
  We consider two different regimes: \emph{nominally}-safe trajectories, which ensure only that the nominal trajectory is collision-free, and \emph{risk}-constrained trajectories optimized using \texttt{ival\_safe} with $\Delta = 10\%$.
  This segregation allows us to identify potentially distinct statistical performance in scenarios representing the full range of risk values $[0,1]$ and over the smaller, ``planning-relevant'' range $[0, \Delta]$.
  These environments are non-convex, and the locally-optimal trajectories were generated via MATLAB's \texttt{fmincon} routine \cite{MatlabOTB}.
  To ensure that the batch of experiments is not dominated by ``uninteresting'' cases, we specifically reject scenarios where the \emph{unconstrained} optimal has negligible failure probability (\ie, when the obstacles are not relevant).
  Computation times for our MATLAB implementation of various methods are shown in Table \ref{table:compute_times} -- as predicted, our method is significantly faster than MC while representing some additional complexity compared to discrete-time methods.
  A more detailed discussion is presented in \refAppRedQuad.

  Some representative scenarios are shown in Fig.~\ref{fig:eval_samples}, and statistics are reported in Table \ref{table:batch_eval} over 100 nominally-safe and 50 risk-constrained scenarios.
  For reference, the average MC-evaluated risk in each of the two test batches is listed in the top of the column.
  The \textbf{Bias} column lists the mean (signed) difference between the estimate and a $1000$-sample MC estimate, and $P(\text{\textbf{Conservative}})$ reports the percentage of cases where the estimate was greater than (or within 5\%) of the MC ``truth.''
  The proposed method \texttt{ival\_safe} outperforms or matches all others in bias, root-mean-squared error (\textbf{RMSE}), and median relative error (\textbf{MRE}), while remaining conservative (\ie, safe) in a significant majority of trials.
  Both \texttt{dt\_gauss} and \texttt{ival\_gauss} outperform \texttt{dt\_booles} but suffer from inconsistency of the Gaussian $\hat{\pi}_k$ estimate.

  \begin{table}[t]
    \centering
    \caption{$\Fz[\x](T)$ estimation statistics for Dubin's car system.}
    \label{table:batch_eval}
    \vspace{-0.3cm}
    \begin{tabular}{l@{\qquad}r@{\quad}r@{\quad}r@{\quad}r}
      \toprule
      \textbf{Batch / Method}         & \textbf{Bias} & \textbf{RMSE} & \textbf{MRE} & $P($\textbf{Conservative}$)$ \\
      \midrule
      Nominally-safe          & MC : 0.2649  & \\
      \texttt{dt\_booles}     &       +2.9780   &  4.908       &    948 \%        & \textbf{100} \%   \\
      \texttt{dt\_gauss}      &       +0.3156   &  0.923       &     70 \%        &  89 \%   \\
      \texttt{ival\_gauss}    &  \textbf{+0.0073}   &  0.257       &     49 \%        &  46 \%   \\
      \texttt{ival\_safe}     &       +0.0915   &  \textbf{0.173} & \textbf{32 \%}   &  92 \%   \\
      \midrule
      Risk-constrained        & MC : 0.0321 & \\
      \texttt{dt\_booles}     &       +0.4849  &  0.957       &   1162 \%      &  \textbf{100} \%  \\
      \texttt{dt\_gauss}      &       +0.0510  &  0.067       &    224 \%      &  96 \%  \\
      \texttt{ival\_gauss}    &       -0.0138  &  0.025       &     \textbf{39} \%      &  30 \%  \\
      \texttt{ival\_safe}     &  \textbf{+0.0057}   &  \textbf{0.019} &   \textbf{41} \%      &  86 \%  \\
      \bottomrule
    \end{tabular}
  \end{table}

\section{Conclusion}\label{sec:conclusion}

This paper addresses the challenging problem of efficiently and accurately estimating failure probabilities to enable risk-aware, continuous motion planning.
By developing a rigorous framework directly in continuous-time and leaning heavily on the classical language of first-exit times, it becomes straightforward to identify a lightweight approximation (\texttt{ival\_safe}) that dramatically outperforms existing methods.
Furthermore, our approximation restores a convenient Bellman structure required for optimal control, enabling practical application for a wide variety of nonlinear motion planning problems.

Ultimately, the framework and concepts introduced in this paper (particularly Section \ref{s:time_additive}) motivate a number of future investigations.
Of particular interest are robust methods of estimating anthropic belief via $\pi_t$ or $\Psi_t$, as ``survival-thus-far'' may represent a useful and yet-uncaptured source of information for aspects of autonomous decision-making not limited to risk estimation.
Also, fusing MC and Gaussian-based risk-approximations in a hybrid approach may provide the best of both worlds: high accuracy in the face of nonlinearity and compatibility with continuous optimization.



\section*{Acknowledgments}
This work was supported by the Education Office at The Charles Stark Draper Laboratory, Inc.\ and by ARL DCIST under Cooperative Agreement Number W911NF-17-2-0181.


\ifdefined\ARXIVVERSION
\else  
  \balance
\fi
\bibliographystyle{plainnat}
\bibliography{IEEEabrv,ref}

\ifdefined\ARXIVVERSION
\clearpage

\newcommand*\zhatt[1][t]{{}^{#1}\!\hat{z}}

\appendix
\subsection{Proofs of Stated Results}\label{app:proofs}
    \textbf{Lemma \ref{lemma:tc_props}}
    \begin{IEEEproof}
      The claims follow almost directly from the definitions; the proof here simply makes this explicit.

      Let $V_{t_1} \triangleq \{s \in [t_1, T] \cbar \z(s) \not \in D\}$ be the set of ``violation times'' after (and possibly including) $t_1$.
      Note that $\tc_{t_1} \geq t_2$ is equivalent to the statement $t_1 \leq t_2 \leq \tc_{t_1} = \inf(V_{t_1})$, and the first claim is immediate.
      The contrapositive is then also established, namely
      \begin{align}
        \tc_{t_1} < t_2 &\iff \exists s \in [t_1, t_2) \text{ s.t. } \z(s) \not \in D  \\
                        &\iff V_{t_1} \cap [t_1, t_2) \neq \emptyset.
      \end{align}

      Moving on to the second claim, note that $V_{t_1} = V_0 \cap [t_1, T]$, where $V_0$ is defined analogously to $V_{t_1}$ above.
      Therefore,
      \begin{equation}
        V_0 \cap [t_1, t_2) = V_0 \cap [t_1, T] \cap [t_1, t_2) = V_{t_1} \cap [t_1, t_2).
      \end{equation}
      Thus,
      \begin{align}
        \tc_0 \geq t_1, \tc_{t_1} < t_2 &\iff \tc_0 \geq t_1 \text{ and } V_{t_1} \cap [t_1, t_2) \neq \emptyset  \\
                                        &\iff \tc_0 \geq t_1 \text{ and } V_0 \cap [t_1, t_2) \neq \emptyset  \\
                                        &\iff \tc_0 \in [t_1, t_2)
      \end{align}
      and we are done.

      The final set of claims follow directly from $z(t)$ \cadlag{} and $D$ closed.
    \end{IEEEproof}
    \vspace{0.5cm}

    \textbf{Proposition \ref{prop:puc_implies_P_conv}}
    \begin{IEEEproof}
      Define the approximating and target events, respectively, as
        \begin{align*}
          &\An \triangleq \big\{\omega \in \Omega \,|\, \max_j \sup_t \zjn(t) \leq 0 \big\} \text{ and } \\
          &E \triangleq \big\{\omega \in \Omega \,|\, \max_j \sup_t z_j(t) \leq 0 \big\}.
        \end{align*}
        Then PUC of $\zn(t)$ to $\vc{z}(t)$ implies that there exist subsets $\Sn \subseteq \Omega$ such that for any $\omega \in \Sn$
        \begin{equation}\label{eq:puc_condition}
          \norm{ \zn(t; \omega) - \vc{z}(t; \omega) } \leq \epsn
        \end{equation}
        and that as $n \rightarrow 0$, $P(\Sn) \rightarrow 1$.
        Define the joint events
        \begin{equation*}
          \Bn \triangleq \An \cap \Sn \text{ and } \En \triangleq E \cap \Sn.
        \end{equation*}
        The additivity of measure over disjoint sets then implies that
        \begin{equation}
          P(\An) = P(\Bn) + P(\An \cap \Snc)
        \end{equation}
        where $\Snc$ is the complement of $\Sn$.
        Because $P(\An \cap \Snc) \leq P(\Snc) \rightarrow 0$, it follows that $\lvert P(\An) - P(\Bn)\rvert \rightarrow 0$.
        An analogous argument establishes that $P(\En) \rightarrow P(E)$.
        Thus, if we can simply show that $\lvert P(\Bn) - P(\En) \rvert \rightarrow 0$, we will be done.

        Again, using the additivity of measure, we have
        \begin{align*}
          &P(\Bn) = P(\Bn \cap \En) + P(\underbrace{\Bn \cap \Enc}_{\triangleq \Cn}) \text{ and }  \\
          &P(\En) = P(\En \cap \Bn) + P(\underbrace{\En \cap \Bnc}_{\triangleq \Dn})
        \end{align*}
        and it will be sufficient to show that the probability of ``disagreement'' $P(\Cn) + P(\Dn) \rightarrow 0$.

        From here, we can consider each of the $m$ elements of the vector processes $\z(s)$ and $\zn(s)$ independently.
        This is because $P(\Cn)$ can be written
        \begin{align}
          &P\big(\Sn \bigwedge \max_j \sup_t \zjn(t) \leq 0 \, \bigwedge \max_j \sup_t z_j(t) > 0 \big)  \\
          &\leq \sum_{l = 1}^m  P\big(\Sn \bigwedge \max_j \sup_t \zjn(t) \leq 0 \, \bigwedge \sup_t z_l(t) > 0 \big)  \label{eq:pt:union_bound}  \\
          &\leq \sum_{l = 1}^m  P\big( \underbrace{ \Sn \bigwedge \sup_t \zjn[l](t) \leq 0 \bigwedge \sup_t z_l(t) > 0 }_{\triangleq \Cn_l}  \big) \label{eq:pt:relax2}
        \end{align}
        where in (\ref{eq:pt:union_bound}) we've used the union bound and in (\ref{eq:pt:relax2}) we've relaxed the probabilistic statement.
        So if we can show that if each term $P(\Cn_j)$ decays to zero (which involves only the $j$-th channel) decays to 0, then $P(\Cn)$ must also decay to 0.
        Furthermore, the PUC criterion for the vector process implies a similar criterion for each of the $j$ elemental processes.
        An analogous decomposition can be performed for $\Dn$, so for the remainder we consider each of the $j$ channels independently.

        Define the random variable $\en \triangleq \sup_t \norm{ \zn(t) - \z(t) }$, allowing $\en = +\infty$ if the supremum does not exist (i.e., within $\Snc$).
        Then $z_j(t) \in [\zjn(t) - \en, \zjn(t) + \en]$ for all $j,t$.

        First, consider set $\Cn_j$, which by construction is a subset of $\Sn$, implying (\ref{eq:puc_condition}) and therefore $\en(\omega) \leq \epsn$ for all $\omega \in \Cn_j$.
        \begin{align}
          P(\Cn_j)  &= P\big(\Sn \bigwedge \sup_t \zjn(t) \leq 0 \bigwedge \sup_t z_j(t) > 0 \big)  \\
                    &\leq P\big( \Sn \bigwedge \sup_t z_j(t) \in (0, \en] \big)  \\
                    &\leq P\big( \sup_t z_j(t) \in (0, \epsn] \big)
        \end{align}
        Because the interval $(0, \epsn]$ approaches the empty set monotonically as $n \rightarrow \infty$, it is clear that $P(C_{n,j}) \rightarrow 0$ for each $j$.

        Similarly for $\Dn_j$, we have
        \begin{align*}
          P(\Dn_j) &= P\big(\Sn \bigwedge \sup_t z_j(t) \leq 0 \bigwedge \sup_t \zjn(t) > 0 \big)  \\
                   &\leq P\big( \Sn \bigwedge \sup_t z_j(t) \in (-\en, 0] \big)  \\
                   &\leq P\big( \sup_t z_j(t) \in (-\epsn, 0] \big)
        \end{align*}
        In this case, the latter probability converges monotonically to $P\big(\sup_t z_j(t) = 0\big)$, which by assumption equals 0.
        Therefore, $P(\Dn)$ also converges to 0.
        Altogether, we have that
        \begin{align*}
          \lvert P(\An)  - P(E) \rvert &\leq \lvert P(\An) - P(\Bn) \rvert  \\
                                       &~~+ \lvert P(\Bn) - P(\En) \rvert  \\
                                       &~~+ \lvert P(\En) - P(E) \rvert \rightarrow 0
        \end{align*}
        completing the proof.
    \end{IEEEproof}
    \vspace{0.5cm}

    \textbf{Proposition \ref{prop:yhatn_puc}}
    \begin{IEEEproof}
      Define the shorthands $\f_t = \f\big( \x(t), \u(t) \big)$, $\G_t = \G\big( \x(t), \u(t) \big)$, $\h_t = \h\big(\x(t)\big)$.

      Consider any $s \in [0,T]$ and the corresponding $t_k \in \pT$ such that $s \in [t_k, t_{k+1})$.
      Then the growth of process $\x(t)$ can be bounded in expectation as follows:
      \begin{align}
        &\E \norm{\x(s) - \x(t_k)} = \E \norm{  \int_{t_k}^s \f_\tau \diff \tau + \int_{t_k}^s \G_\tau \diff \w(\tau) }  \nonumber \\
        &\qquad \leq \int_{t_k}^s \E \norm{ \f_\tau } \diff \tau + \E \norm{ \int_{t_k}^s \G_\tau \diff \w(\tau) } \label{eq:prop1:anal1} \\
        &\qquad \leq \int_{t_k}^s c \diff \tau + \Big( \E \norm{ \int_{t_k}^s \G_\tau \diff \w(\tau) }^2 \Big)^{\frac{1}{2}} \label{eq:prop1:anal2} \\
        &\qquad \leq c (s - t_k) + \Big( \E \int_{t_k}^s \tr \G_\tau^\T \G_\tau \diff \tau \Big)^{\frac{1}{2}} \label{eq:prop1:anal3} \\
        &\qquad = c (s - t_k) + c (s - t_k)^{\frac{1}{2}} \label{eq:prop1:anal4} \\
        &\qquad \leq 2 c (s - t_k)^{\frac{1}{2}}  \label{eq:prop1:anal5}
      \end{align}
      where we appeal to (\ref{eq:prop1:anal1}) the triangle inequality, (\ref{eq:prop1:anal2}) Jensen's inequality, (\ref{eq:prop1:anal3}) the It\^o isometry, and finally (\ref{eq:prop1:anal5}) the fact that $(s - t_k) \leq \delta \leq 1$.

      Due to It\^o's lemma and recalling Assumption \ref{ass:second_order}, the error between $\yhat(s)$ and $\y(s)$ can expressed
      \begin{equation}\label{eq:pf_yhat_error}
        \y(s) - \yhat(s) = \int_t^s \big( \h_\tau - \h_{t_k} \big) \diff \tau.
      \end{equation}
      From here, define the proxy error
      \begin{equation}
        \xi_t(s) \triangleq \int_t^s \norm{ \h_\tau - \h_t } \diff \tau,
      \end{equation}
      which is clearly increasing in $s$.
      From (\ref{eq:pf_yhat_error}) it follows that for any $s \in [t_k, t_{k+1})$
      \begin{equation}\label{eq:xi_inequality}
        \norm{ \y(s) - \yhat(s) } \leq \max_{\tau \in [t_k, s]} \norm{ \y(\tau) - \yhat(\tau) } \leq \xi_{t_k}(s).
      \end{equation}

      By bounding $\xi_{t_k}(s)$ (in expectation) we will be able to bound our error.
      \begin{align}
        &\E \big[ \xi_{t_k}(s) \big] = \int_{t_k}^s \E \norm{ \h\big( \x(\tau) \big) - \h\big( \x(t_k) \big) } \diff \tau  \\
        &\qquad \leq L \int_{t_k}^s \E \norm{ \x(\tau) - \x(t_k) } \diff \tau \\
        &\qquad \leq 2 c L \int_{t_k}^s (\tau - t_k)^{\frac{1}{2}} \diff \tau \\
        &\qquad \leq \frac{4}{3} c L (s - t_k)^{\frac{3}{2}}
      \end{align}

      Recalling that $s - t_k \leq \delta$ by construction, the Markov inequality produces
      \begin{equation}
        P\Big( \xi_{t_k}(s) > \frac{4}{3} c L \delta^r \Big) \leq \frac{ \E \big[ \xi_{t_k}(s) \big] }{\big( \frac{4}{3} c L \delta^r \big) } \leq \delta^{ \frac{3}{2} - r }
      \end{equation}
      for any $r > 0$.
      From (\ref{eq:xi_inequality}) it follows that
      \begin{equation}
        P\Big( \max_{s \in [t_k, t_{k+1})} \norm{ \y(s) - \yhat(s) } > \frac{4}{3} c L \delta^r \Big) \leq \delta^{ \frac{3}{2} - r },
      \end{equation}
      which provides a uniform bound on the probability of excessive error \emph{per segment} in the discretization $\pT$.
      This can be extended via the union bound to compute the probability of excessive error over the full horizon $[0,T]$ as
      \begin{align}
        &P\Big( \max_{t \in [0, T]} \norm{ \y(t) - \yhat(t) } > \frac{4}{3} c L \delta^r \Big)
            \leq \sum_{k = 0}^{K-1} \delta^{ \frac{3}{2} - r }  \\
        &\qquad \leq 2 \frac{T}{\delta} \delta^{ \frac{3}{2} - r } \\
        &\qquad = 2 T \delta^{ \frac{1}{2} - r }
      \end{align}
      where we recall that the number of segments $K \leq 2 T / \delta$.
      Choosing $r = \frac{1}{4}$ produces the desired result
      \begin{equation}
        P\Big( \max_{t \in [0, T]} \norm{ \y(t) - \yhat(t) } > \frac{4}{3} c L \delta^\frac{1}{4} \Big) \leq T \delta^{\frac{1}{4}},
      \end{equation}
      and therefore $\yhat(\cdot)$ converges pathwise-uniformly to $\y(\cdot)$ as $\delta \rightarrow 0$.

    \end{IEEEproof}
    \vspace{0.5cm}

    \textbf{Discussion of Conjecture \ref{conj:jump_probs_negligible}}:

    Each term in the sum (\ref{eq:jump_probs})
    \begin{equation}\label{eq:pf_jump_sum}
      \sum_{k=0}^{K-1} P\big(\tc[\yhat]_0 = t_{k+1}, \yhat_{k+1} \not \in \setOm \big)
    \end{equation}
    can be understood as the probability that the first exit of the approximating process $\yhat(t)$ occurs exactly at $t_{k+1}$, at which point it is outside of the \emph{closed} set $\setOm$.
    This can only occur because $\yhat(t)$, by construction, is discontinuous at each partition point $t_k$, as illustrated in Fig.~\ref{fig:yhat}, and re-sets to match the continuous $\y(t)$ at the start of each $[t_k, t_{k+1})$ interval.
    The event that $\yhat(t)$ has ``jumped out'' of $\setOm$ at $t_{k+1}$ can only occur if $\y(t)$ itself has exited the safe set $\setOm$ within the preceding interval $(t_k, t_{k+1})$ \emph{and} that this ``true'' exit was not reflected in $\yhat$(t).

    The above observation allows us to re-write (\ref{eq:pf_jump_sum}) as the sum of probabilities of such ``undetected'' exits.
    Letting $N$ represent the number of exits of $\y(t)$ in the full horizon $[0,T]$, let $\{U_j\}$ be a collection of indicator random variables.
    Define $U_j$ to be equal to 1 iff the $j$-th true exit of $\y(t)$ went ``undetected'' in $\yhat(t)$ (that is, it produced a ``jump-out`` at $\yhat(t_k)$ for some $t_k$).
    Then it can be seen
    \begin{align}
      &\sum_{k=0}^{K-1} \underbrace{ P\big(\tc_0 = t_{k+1}, \yhat_{k+1} \not \in \setOm \big) }_{ \text{prob.\ first exit is a jump-out at $t_{k+1}$.} }  \\
      &\leq \sum_{k=0}^{K-1} \underbrace{ P\big(\tc_k = t_{k+1}, \yhat_{k+1} \not \in \setOm \big) }_{ \text{prob.\ of jump-out after $(t_k, t_{k+1})$} }  \\
      &= \E \Big[ \sum_{k=0}^{K-1} \underbrace{ U( t_k, t_{k+1} ) }_{\text{indicator that a jump-out occurs at $t_{k+1}$}} \Big] \\
      &\leq \E \Big[ \sum_{j = 1}^N U_j \Big]  \label{eq:pf_exp_Uj}
    \end{align}

    Crucially, $N$ (the number of true exits) is a characteristic of the true process $\y(t)$, and is thus independent of $\pT$ and $\yhat$.
    Because each $U_j \in \{0, 1\}$, the first claim that (\ref{eq:pf_jump_sum}) is upper-bounded by $\E [N]$ follows immediately.

    The $U_j$ indicators depend on $\pT$, and though it is difficult to prove, it seems likely that as $\pT$ is refined, each $E[U_j] \rightarrow 0$, due to the pathwise-uniform convergence of $\yhat(t)$ to $\y(t)$.
    Because $N$ is independent of $\pT$, this elementwise decay to zero is sufficient to ensure that the sum (\ref{eq:pf_exp_Uj}) also decays to zero.
    \vspace{0.5cm}

\subsection{Reduced Quadratures: Second-Order Gaussian Systems}\label{app:red_quad}
  \balance

  We would like to evaluate the following $n_x$-dimensioned integral for each constraint $j$
  \begin{align}
    \Phiz[\yjhat](t_k, t_{k+1}; \mu_k) = \int_{\manX} \Ind\big(\tc[\yjhat]_k < t_{k+1} \cbar \x_k = \x \big) \diff \mu_k(\x) \label{eq:quad_full}
  \end{align}
  which is most naturally taken over the state space $\manX$, as we usually are operating from state-space distributions $\pi_k$ or (non-normalized) $\Psi_k b_k$.
  Indeed, we will specifically assume that $\mu_k$ represents either $\Psi_k b_k$ or the conservative $\Ind_{\Xsafe} b_k$, and thus has no support over $\cXsafe$.
  Moreover, because $\yjhat$ specifically follows a constant-coefficient, linear form over the interval $[t_k, t_{k+1}]$, we can write (\ref{eq:quad_full}) as
  \begin{align}
    &\int_{\manX} \Ind\big(\yjhat(t_k) + h_j(t_k) \delta_k > 0 \cbar \x_k = \x \big) \diff \mu_k(\x)    \\
    &= \int_{\manX} \Ind\big(g_j(\x) + h_j(\x) \delta_k > 0 \big) \diff \mu_k(\x)  \label{eq:quad_full_specific}
  \end{align}
  For the purpose of comparison, we point out that discrete-time direct methods based on (\ref{eq:dt_booles}) require evaluation of a similar integral
  \begin{equation}\label{eq:P_violation}
    P\big( \x(t_k) \in \cXsafe \big) = \int_\manX \Ind(\x \in \cXsafe) \diff \mu_k(\x).
  \end{equation}

  The evaluation of such multi-dimensional integrals in general, and in the setting assumed by this paper, must be performed numerically, often via quadrature methods.
  The accuracy and efficiency of such methods is highly sensitive to dimensionality of the integration, and thus practical realtime implementation requires dimension $\leq 3$.

  Consider the common example of a second-order physical system, which evolves as a nonlinear ``integrator''.
  That is, the state space can be partitioned as $\x = (\p, \vc{v}, \ldots)$, such that the position dynamics simply follow a velocity state
  \begin{equation}\label{eq:p_dynamics}
    \vcdot{p}(t) = \vc{v}(t).
  \end{equation}
  Combined with the condition that the constraints are functions of this position substate $\vc{g}(\x) = \vc{g}(\p)$, such a system clearly satisfies the conditions of Assumption \ref{ass:second_order}.
  This state structure allows us to write $h_j(\p, \vc{v}) = \vc{a}_j(\p)^\T \vc{v}$ and therefore (\ref{eq:quad_full_specific}) as the nested integral
  \begin{align}
    &\Phiz[\yjhat](t_k, t_{k+1}; \mu_k) =  \label{eq:quad_ph} \\
    &\, \int_\Rnp\!\!\!\! \Ind(\p)_{\Xsafe} \int_\Rnp \!\!\!\! \Ind\big( g_j(\p) + h_j \delta_k > 0 \big) \diff b_k\big(h_j \cbar \p\big) \diff b_k(\p)  \nonumber
  \end{align}
  where the with a slight abuse of notation we've used $b_k(h_j \cbar \p)$ and $b_k(\p)$ to represent the conditional and marginal distributions, respectively.

  In the common case that $b_k$ follows a multivariate Gaussian distribution, the conditional and marginal distributions of $h_j(\p, \vc{v})$ and $\p$ are straightforward to compute exactly.
  Moreover, we can simplify the inner integral to produce
  \begin{equation}\label{eq:quad_p}
    \int_\Rnp\!\!\!\! \Ind(\p)_{\Xsafe} \phi\Big(\frac{g_j(\p)}{\delta_k}; -\bar{h}_j(\p), \sigma_j(\p) \Big) \diff b_k(\p)
  \end{equation}
  where $\bar{h}_j(\p)$, $\sigma_j(\p)$, and $\phi$ refer to the mean, standard deviation, and CDF of the one-dimensional normal distribution $b_k(h_j \cbar \p)$.
  Given that the function $\phi$ is available directly in most computing environments, (\ref{eq:quad_p}) now represents a relatively compact $n_p$-dimensional quadrature.


  %

\subsection{Simulated Dubin's System}\label{app:sys}
  Our experimental validation relies on a simulated second-order Dubin's car with state $\x = (\p, \vc{v}, \theta, \omega) \in \R^6$, and control $\u = (c, \alpha)$.
  $\theta$ and $\omega$ capture the heading angle and angular rate, respectively, while $c$ refers to forward-pointing thrust and $\alpha$ the angular acceleration.
  The time-invariant, nonlinear dynamics are given by
  \begin{equation}
    \vc{f}(\x, \u) = \begin{bmatrix} \dot{\p} \\ \dot{\vc{v}} \\ \dot{\theta} \\ \dot{\omega} \end{bmatrix}
                   = \begin{bmatrix} \vc{v} \\ c \mx{R}(\theta) \vc{e}_1 \\ \omega \\ \alpha \end{bmatrix}
  \end{equation}
  with constant noise matrix
  \begin{equation}
    \mx{G} = 0.05 \begin{bmatrix} \mx{0} & 0 & 0 \\
                                  \mx{I}_2 & 0  & 0 \\
                                  \mx{0}  & 0.1 & 0 \\
                                  \mx{0}  &  0  & 1 \end{bmatrix}.
  \end{equation}

  To simulate output-feedback control, we introduce a simple observation process at each timestep $k$
  \begin{equation}
    \vc{y}_k  = \x_k + \vc{\nu}
  \end{equation}
  where $\vc{\nu} \sim \mc{N}(0, 0.0001 \mx{I}_6)$.
  Then a corresponding LQG-style controller is constructed to track a nominal trajectory that also defines the linearization point.

\fi

\end{document}